# A Distributed Reinforcement Learning Solution With Knowledge Transfer Capability for A Bike Rebalancing Problem

Ian Xiao | New York University | ian.xxiao@nyu.edu | LinkedIn

## Abstract

Rebalancing is a critical service bottleneck for many transportation services, such as Citi Bike. Citi Bike relies on manual orchestrations of rebalancing bikes between dispatchers and field agents. Motivated by such problem and the lack of Reinforcement Learning (RL) application in this area, this project explored a new RL architecture called Distributed RL (DiRL) with Transfer Learning (TL) capability. The DiRL solution is adaptive to changing traffic dynamics when keeping bike stock under control at the minimum cost. DiRL achieved a 350% improvement in bike rebalancing autonomously and TL offered a 62.4% performance boost in managing an entire bike network. Lastly, a field trip to the dispatch office of Chariot, a ride sharing service, provided insights to overcome challenges of deploying an RL solution in the real world.

## Acknowledgement

I want to thank the NYU CUSP faculty to provide the freedom for me to explore this topic. Special thanks to Prof. Stanislav Sobolevsky and Sattik Deb for their guidance in shaping the project scope and direction. At the very beginning, I wanted this project to strike a balance between innovation and practicality. Thanks to Saad Shaikh, Peter Lau, and Saul Garza from Chariot, who allowed me to visit their field operation, I was able to ground this project to reality by understanding what actually happens in a dispatch room and the business priorities. Finally, the insights and advice about putting AI systems in operation from Kris Sandor, the General Manager of Citi Bike, were priceless.

## Code

All the codes developed and used in this project can be found here in my Github Repository.

**Key Terminology**

This report involves an extensive list of terminologies that can be overwhelming to readers who are less familiar with Reinforcement Learning. The following list hopes to provide some context.

- **Reinforcement Learning (RL)**: a machine learning paradigm that allows a program to improve its action through trial-and-error from interaction with the environment.
- **Distributed Reinforcement Learning (DiRL)**: an RL design that focuses on breaking the system down to the most fundamental and independent elements and apply RL in a parallelized fashion.
- **Agent**: Reinforcement Learning object acting as a "bike re-balancing operator".
- **Policy**: agent's behavior function, which is a map from state to action
- **Value Function**: a prediction of future rewards
- **Model**: agent's representation of the environment
- **Transfer Learning**: a solution that allows agents to share knowledge across domains and time.
- **Naive Agent**: an agent that does not have any prior knowledge received from Transfer Learning.
- **Experienced Agent**: an agent that has knowledge from prior training received from Transfer Learning.
- **Environment**: a bike station object that will provide feedback such as the number of bikes and reward or penalty.
- **State**: the number of bike stock at a given time (e.g. 23 bikes at a station in hour 3).
- **Training**: interactions between the agent and environment for the agent to learn what the goal is and how to achieve it the best.
- **Episode**: number of the independent training session (the environment is reset, but agent keeps the learning from one episode to another); each episode has 24-hour inter-dependent instances with bike stock info based on the environment setup and agent actions

- **Session**: each session has multiple episodes with both environment and agent reset; the goal is to benchmark agent performances based on the number of episodes (e.g. will more training episode leads to high success ratio? When should we stop the training?)
- **Q-Table**: a matrix the agent use to decide future action based on state-action-reward tuples; the agent develop this Q-Table from each training episode based on environment feedback.

## 1. Introduction

Rebalancing problem remains one of the most critical issues to sustain and expand operations for transportation businesses. This issue is even more prominent for businesses that operate in an environment with asymmetric commuting patterns, such as bike sharing. According to the monthly operating report, Citi Bike spends a significant amount of effort on rebalancing bikes every day (Barone, 2017). This is a substantial problem to solve. There are two main methods to rebalance today: providing an incentive to users to help move bikes (e.g. the Bike Angel program) and using a trailer operated by field agents across the city. Managing a fleet of vans, back-office staff, and field operators are arguably some of the largest cost drivers.

Using an autonomous and intelligent approach enabled by Distributed Reinforcement Learning (DiRL) can translate into cost-saving opportunity in two ways: 1) better optimization of the resource (e.g. frequency of trips, number of bikes moved per trip, etc.) with a realistic and timely representation of the system dynamics, and 2) more robust model because of the solution's continuous adaptability to changing system dynamics.

This project contributes to the rebalancing problem in the following ways. First, this project is one of the very first solutions that use RL to tackle large-scale rebalancing issues. Second, from a technical standpoint, this project experiments and highlights key benefits and shortfalls of using distributed RL scheme and Transfer Learning. Lastly, I used this project as an opportunity to interview front line staffs in order to understand how AI can improve their daily work.

## 2. Literature Review

In short, the Literature Review helps to highlight the opportunity in applying RL and DiRL to solve the Citi Bike Rebalancing problem because 1) rebalancing problem has been tackled by using other traditional machine learning techniques with limitations, 2) RL is typically applied to gaming, finance, and physical control systems, but not to asset allocation problem like the bike rebalancing issue, which will require a new architecture to handle the scale, and 3) the applications of RL, especially with an ability to transfer skill through shared experience, is one of the most important AI research topics. The following paragraphs elaborate on each factor.

**Rebalancing problem is typically tackled with traditional machine learning techniques, which requires expensive retuning when deploying in different contexts**. According to the report on Citi Bike, operators at the 24-hour dispatch center are relying on a digital station map and forecasting model that takes in weather information. Instructions are sent teams in the field (Barone, 2017). In parallel, the academic communities have been looking into rebalancing problems in not only bike sharing, but many other transportation domains. Various methodologies and use of data were proposed and studied. For example, in An Intelligent Bike-Sharing Rebalancing System (Lopes, 2015), Diogo Lopes investigates an array of machine learning techniques for predicting how bikes will move throughout bike-share systems on an hourly basis. Forming training and test sets by splitting data gathered from Washington D.C. and Chicago bike-shares, Lopes examines Bayesian Networks, Extra Trees, Gradient Boosting Machines, Linear Regression, and Poisson Regression as potential algorithms for predicting bike-movement throughout the network; he graphs error-rates for each algorithm over 1-, 2-, and 3-hour intervals from given start-times. In each of these three-hourly intervals, Gradient Boosting Machines prove to be the most effective at predicting bike movement. However, there is a noticeable upward shift in error rate with each additional hour.

Historically, rebalancing is often viewed as route optimization problems because it involves human traveling within cities physically. Under this setup, it is commonly solved with multi-traveling salesman formulation based on the work by Mehdi and his team (Nourinejad,

2015) In a recent paper, Jasper and his team proposed to solve the rebalancing problem with a mixed method by combining inventory prediction and route optimization (Schuijbroek). In terms of shortfalls, many of the methods mentioned assume data availability and depend on a sophisticated modeling of the stochastic behavior of the traffic dynamics, which can be challenging in the real world. In addition, traditional models would require extensive re-coding and validation if one wants to take the existing model and apply to new locations and use with new data.

This reveals an opportunity to find a solution that relies on common operational data and can be easily re-tuned with the less ongoing effort from expert modelers. RL is a good fit because of its ability to learn and improve autonomously in complex and changing environments.

**RL is typically applied to games, finance, digital marketing and robotics, but not to urban operation problems.** Since its inception in the 1950s, RL has proven its effectiveness in closed-loop systems, such as games, finance, and physical control systems (e.g. room temperature control, manufacturing, and autopiloting). Deep RL (DRL) using neural network techniques, a branch of RL, had attracted tremendous research and commercialization investment because of its power in storing and processing high dimensional parameters and improve autonomously (Kai Arulkumaran, 2017). Essentially, this removes the computational bottleneck in creating general artificial intelligence. In the famous publications on Nature journal, David Silver, Volodymyr Mnih, and their teams demonstrated how DRL helped to train a computer program to play various games without human knowledge and intervention (Mnih, 2015) (Silver, 2017). In 2017, a computer program called AlphaGo, which learned to play the game of Go using DRL, beat the best human players in the world (Etherington, 2017). Beyond gaming, Investment Banks and Hedge Fund started to use DRL to gain a competitive advantage both in practice and in research. JP Morgan reported that they deployed a DRL enabled solution, called LOXM, to execute equity trades to maximize speed and at optimal prices (Terekhova, 2017). In China, Zhengyao Jiang and his team published their solution of applying DRL to manage financial portfolios. Many of the traditional techniques, such as Q-Learning, Deep Q-Learning

using Convolutional Neural Network (CNN) and Long Short-Term Memory (LSTM), were implemented and benchmarked (Jiang, 2017). DRL also found its home in digital marketing. A team of researchers at JingDong.com, which is the largest e-commerce platform in China, published their work on combining DRL with traditional recommender systems. By doing so, their solution was able to perform better offline learning, manage extremely large product and user parameters, and achieve higher marketing response based on tests in the real-world (Zhao). The Google Brain and Boston Dynamics (a subsidiary of Google) have been investing heavily in fundamental RL research and its application to robotics. The specifics will be discussed in the next paragraph.

**RL (or DRL) has not yet been extensively applied to operation problems based on the literature in both academic and corporate R&D communities.** Rebalancing shares many environmental characteristics with gaming, finance, and marketing. For example, the environment is complex and high dimensional, the system is closed-loop, and the underlying system drivers (e.g. economics condition and customer preferences) are constantly changing. The drivers that change the system will be the traveling pattern and new geographic constraint in the case of rebalancing. That said, applying RL to rebalancing can be a valuable attempt to push the boundary and mitigate limitations of current approaches, such as inflexibility, data requirements, and complex feature engineering.

More importantly, RL has yet to prove viable in big problems where the system dynamics are large and nuanced and the agent has a high degree of freedom in choosing actions. I believe the main reasons are the limited computation power - Google Alpha Go as an exception because of Google's computation resources. Many companies do not have the computation luxury that Google has and it is not realistic to invest in high performing computing infrastructure immediately without a solid business case. Therefore, scaling RL solutions requires a different solution - an architectural one. Using existing resources better with parallel computing is a viable option as we have seen in the recent Big Data evolution. The distributed design scheme can be applied to RL to allow the agents to learn faster and act better. This motivates the experimentation of Distributed RL (DiRL) in this project.

**Pushing the boundary of RL application in another dimension by adopting a new architecture and Transfer Learning.** Finding ways to reduce the learning time and increase the performance of DRL is a critical research topic at the moment. A popular solution is by incorporating skill transfer through collective learning and knowledge representation. In 2016, a team at Google Brain implemented a technique called Collective Robot RL with Distributed Asynchronous Guided Policy Search to prove the effectiveness in reducing training time and performance (Ali Yahya) (Levine). In February 2018, a team of scholars from the University of Southern California published a paper on the similar topic, but proposed a probabilistic approach in order to generalize the experience sharing (Hausman, 2018). In all the research work mentioned, finding a good way to represent knowledge, transfer this knowledge from an agent to another or to a new problem domain, and allowing the machine to learn by imitating human action remains one of the most active research investments.

A general architecture of knowledge accumulation, distillation, and transfer has been proposed by a team from the University of Sao Paulo at the Artificial Intelligence conference (Silva, 2016). The architecture (Figure 1) involves a knowledge repository that allows teacher agents and new agents to upload, download, and update knowledge. However, the team did not show an implementation and concrete results. Implementing a similar TL architecture is an opportunity for this project.

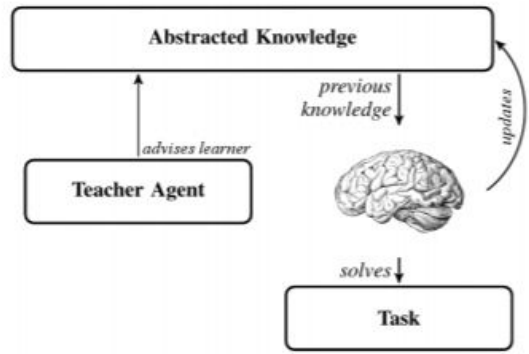

**Figure 1**: A Conceptual Architecture Diagram of Multi-Agent Transfer Learning (Silva, 2016)

Robotics is an area with tremendous innovation in TL, especially in a field called Imitation Learning. A team of researchers at UC Berkeley, led by Prof. Levine, has produced a large body of work around Imitation Learning to allow machines to learn from human demonstrations. A recent paper illustrated a solution to teach machines how to pick up objects and move them to a changing target drop zone (Rahmatizadeh, 2018). This requires machines to find ways to represent and generalize the observed knowledge captured by image sensors. More importantly, the machine can then transfer the learning to a different environmental setup, such as a new object specification and target drop zone.

Although the TL solutions discussed above are not applied to asset rebalancing context, the main idea of representing and sharing knowledge from multi-agent learning can be adapted to the DiRL solution envisioned in this project.

To motivate this project, the literature review highlights opportunities to apply RL to a large-scale operation problem such as bike rebalancing, experiment a new distributed learning and computation architecture, and incorporate Transfer Learning to improve the learning performance. With this in mind, the following section specifies and quantifies the bike rebalancing problem to measure the performance of the proposed DiRL solution.

## 3. Problem Definition

One of the key performance metrics for Citi Bike is bike availability at each station throughout the day. With this in mind, the most important objective for the DiRL solution is to maintain the bike stock of each station in a network to be within a fixed range - this is defined as the Success Ratio. This translates to the following measurement:

$$\text{Successful Bike Station} = \begin{cases} 1 & \text{if each hour's bike stock is between 0} \\ & \text{and initial stock*threshold, inclusive} \\ 0 & \text{if otherwise} \end{cases}$$

Secondly, to be as realistic as it can be in a business setting, the DiRL solution should achieve this with the minimum cost. The following reward and penalty structure help to do so:

$$\text{Reward / Penalty} = \begin{cases} 50 & \text{if bike stock is within the range at hour 23} \\ -50 & \text{if bike stock is outside the range at hour 23} \\ -30 & \text{if bike stock is outside the range at any hour before hour 23} \\ -0.5 & \text{for every bike moved in each hour} \end{cases}$$

The reward or penalty is given to each agent in the DiRL network at each hour. At the end of each episode, a Total Reward, which is a summation of the net reward (e.g. reward - penalty) of all agents, is calculated and tracked for benchmarking. All in all, the agents are set up to maximize rewards they collect over the session.

Lastly, based on common practice in TL (Lazaric, 2013), the knowledge transfer improvement is measured by the difference between the Areas under the Reward Curve, which is denoted as R with the following definition:

$$R = \frac{\text{Area with Transfer - Area without Transfer}}{\text{Area without Transfer}}$$

## 4. Methodology

The methodology section focuses on providing an overview and detailed breakdown of the solution and mathematical approach of the knowledge distillation process in Transfer Learning.

### 4.1 A Conceptual Diagram of the Distributed RL Solution

Figure 2 illustrates the station-agent pairs and interactions with a Knowledge Repository. The Distributed RL design features a parallelized interaction and learning scheme. Each agent learns and acts independently while the learnings are curated and shared via the Knowledge Repository. The distributed RL focuses on breaking down the problem to the most fundamental level, which is different from the common reference to breaking down computation to multiple threads. The specific mechanics of how knowledge transfer and sharing will be discussed in sections below.

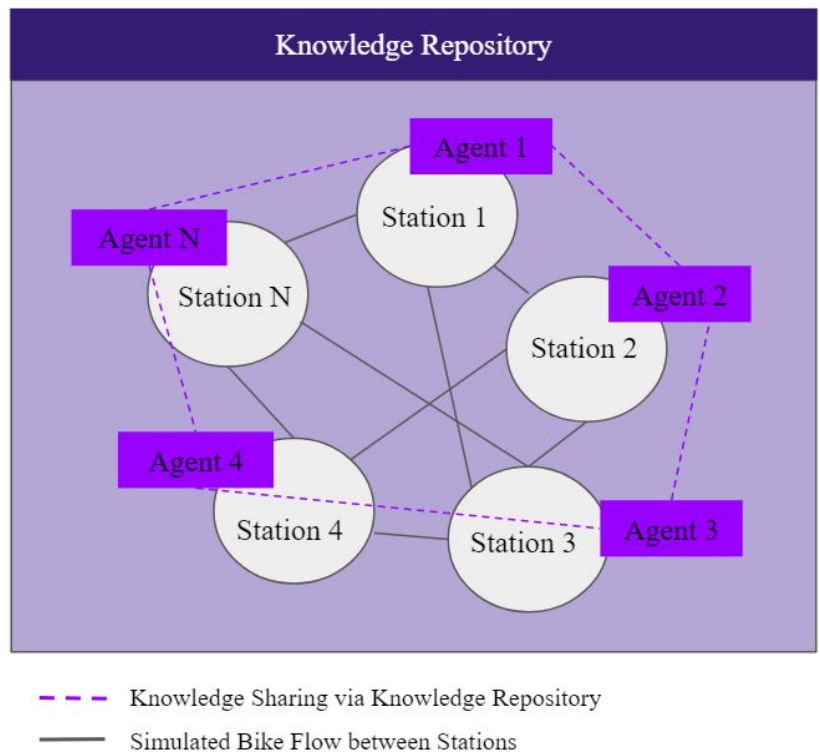

**Figure 2**: A High-Level Diagram of Distributed RL solution

One may wonder why a distributed design was chosen, instead of having a single RL agent managing the whole network. The following two sections will address this question.

## 4.2 Advantage and Disadvantage of Having A Single RL Agent Overseeing a Bike Network

Having a meta-RL agent to learn and manage a bike network is an intuitive solution. However, having to handle an exponentially increasing action space as the bike network expands is a critical downside. For example, an action space can be defined as the number of bikes the agent can move between two stations. For a network of 3 stations, the RL agent can move 1, 3, or 5 bikes from station 1 to 2, from station 2 to 1, from station 1 to 3, and from station 2 to 3, and so forth. The total number of viable actions, A, can be calculated as the following:

A = Number of Viable Actions * Number of Edges in the Bike Network

Where the number of edges in a network increases exponentially with the number of stations, n:

$$\text{Number of Edges in a Network} = \frac{n(n-1)}{2}$$

Many existing RL solutions are designed to handle large state space using, for example, a Convolution Neural Net, but most solutions only learn to find optimal solutions in a limited action space (e.g. move up, down, left, right in games).

## 4.3 Advantage and Disadvantage of Having a Distributed RL Scheme

The distributed RL scheme is designed to avoid having to handle an exponentially expanding action space. Having station-agent pairs is essentially taking a bottom-up approach, which allows each agent to find the optimal solution for each bike station and aggregate subsequently. The action space for each agent is fixed regardless of the size of the network. For instance, the size of action space is only 3 because the agent only needs to decide if it should move 1, 3, or 5 bikes from the station it is managing. In this case, the number of RL agent has to increase, but it only increases linearly based on the size of the network. Relatively speaking, the distributed RL design may have better scalability.

There are two key disadvantages of having a distributed RL scheme. First, an extra layer of logic is required to consolidate the decisions for the system; however, I think the computation overhead introduced by this extra layer is reasonable. Secondly, it demands a higher run-time processing power to support the parallel learning of all RL agents. Regardless, I think the trade-off for a more reliable and scalable solution is worth it.

This is a high-level overview of how the Distributed RL solution works. The following sections dive into the details of the key components: station-agent pair and transfer learning using the Knowledge Repository.

### 4.4 The Station-Agent Pair Architecture

The station-agent pair interactions are the most fundamental elements of the solution. Shown in Figure 3, each station-agent pair can be broken down to interactions between a station environment and an RL agent. All pairs are synced by a common 24-hour clock for all station-agent pairs in a network.

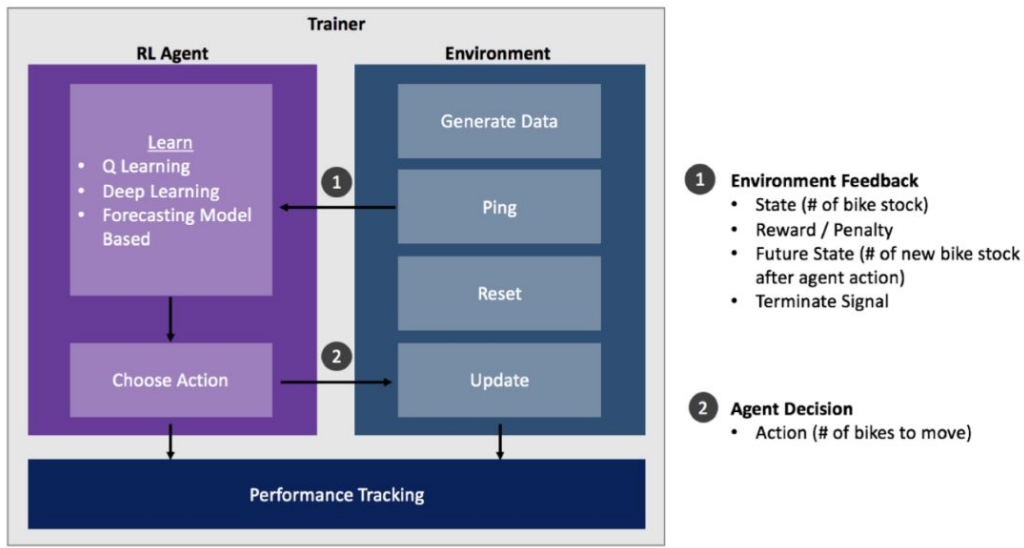

**Figure 3**: An Architecture Diagram of Station-Agent Pair

**A Station Environment** contains the following major functions.
- **Generate Data**: simulate hourly bike stock flow that coordinate with other stations in the network

- **Ping**: receive instruction about how many bikes to move from the paired RL agent; send feedback, such as old bike stock, updated bike stock based on the instruction from the RL agent, and reward/penalty based on the updated bike stock (more details below)
- **Update**: calculate new bike stock based on RL agent instruction and propagate the changes to future hours
- **Reset**: re-initiate the station environment for a new episode by creating new simulated bike flow while keeping between-episode analytical data for performance benchmarking and debugging

**An RL agent** includes the following properties and functions to provide learning, decisioning, and collaboration capabilities.
- **Learn**: use Q-Learning to create a Q Table of state, action, and value mapping based on reward/penalty feedback from the Station Environment.
- **Choose Action**: pick the optimal action based on the current bike stock and Q-Table
- **Upload**: send learning to the Knowledge Repository
- **Download**: receive insights from the Knowledge Repository to assist with decision making. It uses a weighting scheme to consolidate the agent's own knowledge and the collaborating agents.

Lastly, a **Performance Tracking** module tracks key indicators, such as an individual station success rate, a collective success rate in a network, individual and collective rewards collected by the agent.

**4.5 Transfer Learning Mechanism and Measurement**

The goal of Transfer Learning (TL) is to shorten the learning time to achieve high performance (Lazaric, 2013). A Knowledge Repository (KR) was created to facilitate the TL process. The KR collects learnings from all agents and distills them into a single Q-Table for future sharing. In this solution, TL manifested in two distinct ways: in-session and between-session knowledge transfers. Figure 4 illustrates the Transfer Learning process.

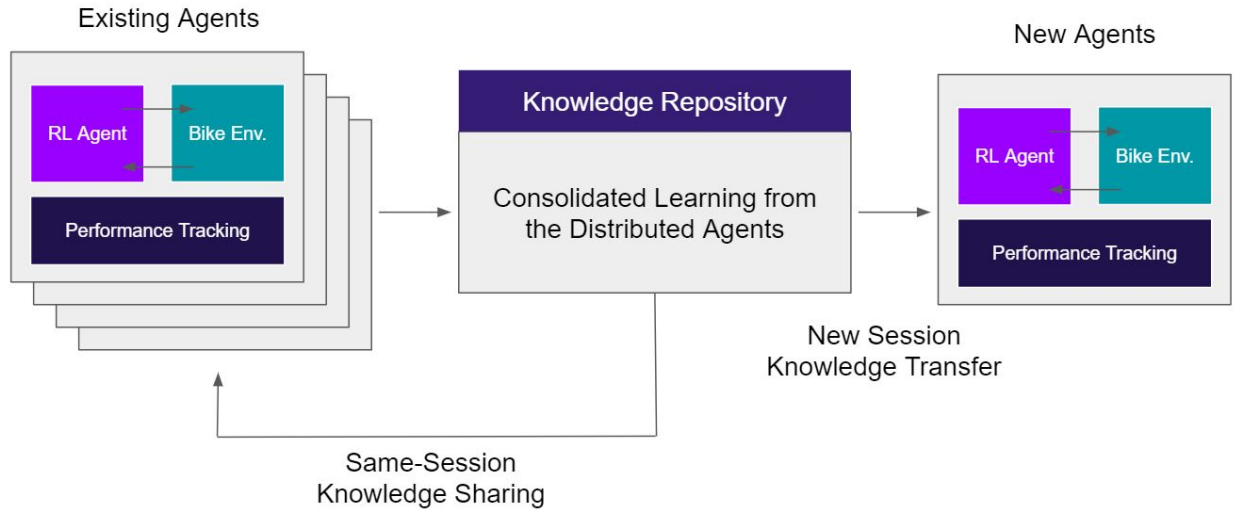

**Figure 4**: A Process Diagram of Transfer Learning

In practice, same-session knowledge transfer can mean sharing learnings amongst agents in New York City in real-time or between days. Between-session knowledge transfer means translating the learnings from NYC to new agents, for example, in Seattle.

As mentioned in the Problem Definition section, a common measurement of TL performance is to compare the areas under the reward curve over all the episodes. In general, there are three kinds of improvements demonstrated by Figure 5: learning speed, asymptotic (long-term success), and jumpstart improvements. Each improvement offers unique benefits depending on the problem. The jumpstart improvement signifies both speed and early stage performance boosts.

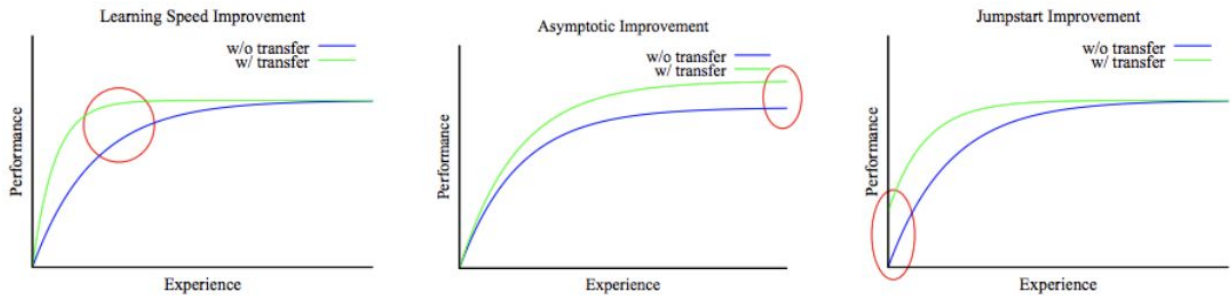

**Figure 5**: Three Types of TL Improvements (Lazaric, 2013)

## 5. Results & Discussion

This section discusses the key results, limitation, and future work of the DiRL solution. The key is to highlight the working components, improvement areas, and ideas for better solutioning.

### 5.1 The Key Results: Bike Network Rebalancing with Transfer Learning

**Overall Performance.** The key metrics the DiRL tracks are total rewards and overall success ratio. According to Figure 6, the DiRL is able to improve both metrics as the solution interact and learn from more episodes autonomously. Mostly importantly, the Success Ratio, defined as the number of successfully rebalanced station over the total station in a network, increased from about 10% to 35% by the end of the training session with 100,000 episodes.

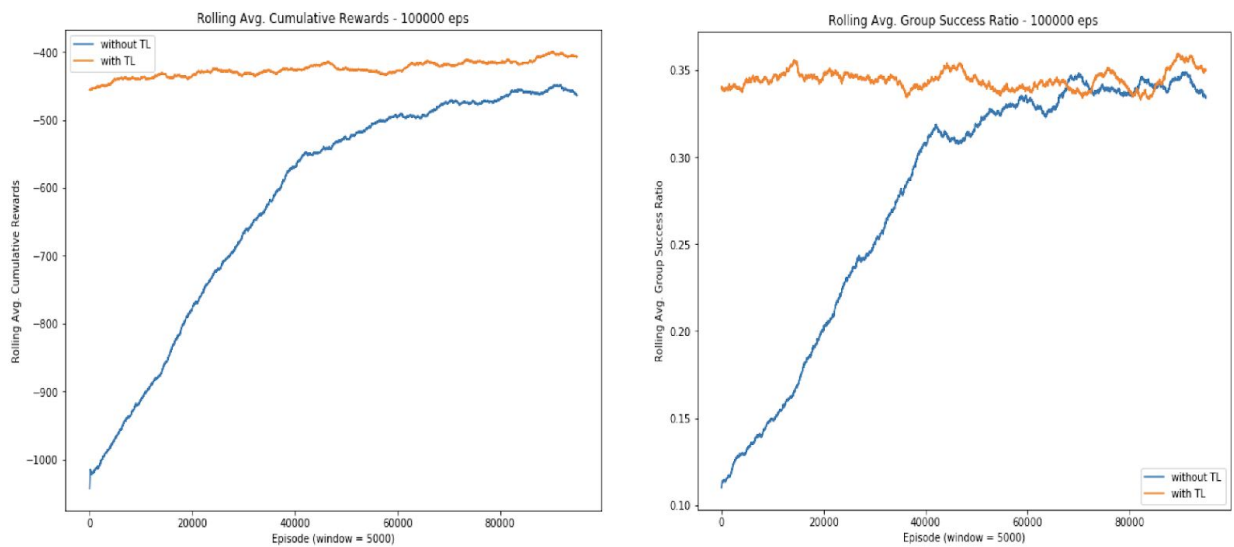

**Figure 6**: Benchmarking data on the Total Reward (left) and Success Ratio (right); blue and orange lines represent results of bike balancing without and with TL respectively

**TL Performance**. In addition to the overall rewards and success ratio, the count of complete network success (e.g. all stations were within range throughout the whole day) provides an intuitive benchmarking on the impact of TL. Shown in Figure 7, the DiRL with the TL capability was able to completely rebalance a network 62.4% better than the one without TL.

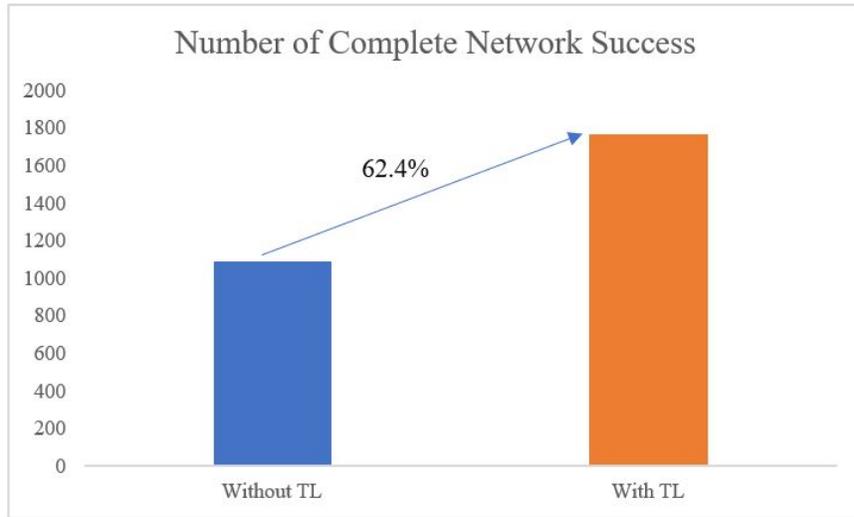

Note:
- total number of episodes: 100,000
- action allowed: [-30, -20, -10, -3, -1, 0 1, 3, 10, 20 , 30]

**Figure 7**: Comparison of Complete Network Success (all stations are within limits in all hours)

**Cost of Operation.** Furthermore, the DiRL was able to achieve the overall rebalancing improvement with a progressively lowering cost. Based on Figure 8, the cost of moving bikes decreases as the DiRL learned from additional episodes; and it eventually managed to reduce the cost by 15%. This means that the DiRL was able to rebalance the network by moving less or just enough bikes.

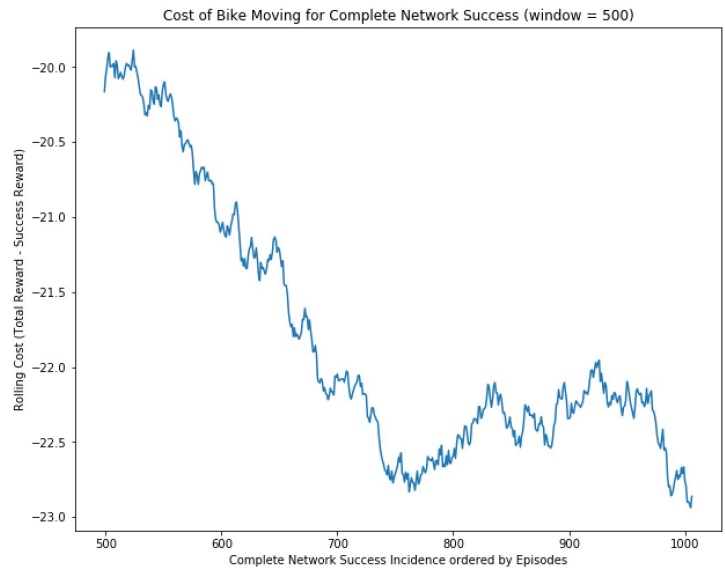

**Figure 8:** Cost of moving bike when complete network success was achieved

However, the DiRL with TL was only able to completely rebalancing about 2% of the training episodes. The following section will discuss the potential cause and workarounds to improve the overall performance.

### 5.2 Key Characteristics and Limitations

Figure 6 and 7 illustrate what the DiRL was able to achieve, but also lacking. The following discuss the three main observations in more details.

**Distributed RL works.** One can see the Total Reward (the blue line) increases as all the agents interact with their corresponding bike environments. This proves that a distributed RL scheme works similarly to a theoretical single-RL setup. This is because the DiRL solution breaks the system down to its most fundamental working blocks that share the same dynamics. If the solution works at the fundamental level independently, it will work at a system level.

**TL helps to jumpstart the performance, but only in the beginning**. The TL helps to create about 23% improvement using the R metric defined in the Problem Definition section. The contribution mainly comes from early episodes when the naive agents still lack the knowledge compared to the experienced agents. The experienced agents were able to perform well at the get-go, which is signified by the high reward in early episodes. However, the learning improvement for the experienced agents stays fairly constant over all the episodes. Intuitively, a reason can be that the DiRL solution already exhausted all learning potential, which means the new knowledge can be discovered through additional interactions is minimum.

**Better reward translates to high success, but with some limitations.** In general, the Success Ratio increased as reward improved through learning and Transfer Learning. The question is that why the average success ratio is limited around 35% and only 2% for complete network success. Arguably, such success rate is not very high compared to any modern industrial solution. There is a number of explanations. First, the bike dynamic was simulated based on the random hourly flow between -20 and 20. In some cases, this flow may create unsolvable scenarios. Despite the agents executed based on the best strategy from learning, they simply could not keep the bike stock under the limit given the number of actions they are allowed to take (e.g. only choose to move 1, 3, or 5 bikes while 20 bikes are deposited). Secondly, the test case only included 3 stations due to computation limit of the hardware. The average success ratio was very sensitive. For example, having 2 out of 3 stations being successful reduce the ratio from 100% to only 66%. Theoretically, the Success Ratio will be more stable when managing a bigger network of bike stations. Lastly, the definition of success was very strict, which was designed on purpose to be conservative. For example, one can simply increase the threshold of "overstock" from 1.2 to 3 to allow more room of error or only define success based on the stock in the final hour instead of throughout the day.

## 5.3 Future Work

Due to time constraints and priority, there are many features did not get to be implemented and some solution designs can be further polished. The following highlights some key ideas for immediate improvements.

**Increase in computing performance.** Expanding the size of DiRL to cover a larger network requires exponential computation power. This is critical to the scaling of the solution. There are three immediate solutions. First, improve the hardware set up by running the solution to a cloud-based infrastructure with cheap, but scalable, hardware. Amazon and Google Cloud are two of the best options given their pricing and easy-to-maintain set up. The second option is to enable GPU computing on a desktop machine. The last option is to have low-level algorithm optimization. For example, one can enable multi-CPU computing using a python library called MultiThreading. In addition, a more efficient data retrieval process for the Knowledge distillation and transfer can be developed. The current solution uses a crude solution: the agents deposit their knowledge every 100 episodes.

**Unlocking rebalancing performance**. This is the most important metric in proving the value of the solution. Theoretically, a substantial amount of performance can be unlocked via better hardware because the hardware allows more and faster learning. Secondly, a more realistic simulation can help to bound the possible bike variation throughout the day. The current solution uses a fixed range based on an arbitrary selection. Lastly, a new scheme of autonomous and adaptive action space modification by the DiRL can ensure the agents has enough "freedom of movement" to manage the bike variation without extensive up-front tuning. For example, the DiRL can expand its action space from [-3, 0, 3], which will be designed by a human programmer, to [-10, -5, -3, 0, 3, 5, 10] on its own based on learning. This is, I believe, the line between human controlled and fully autonomous AI.

**Design a generalizable interface.** Being able to deploy the DiRL solution to different problem domains without extensive re-coding is a key to real-world deployment. Different problem domains require the DiRL to take in different input data and data storage scheme. For example, having a DiRL to automatically reroute fleets may require a different way to digest and store the information of fleet positions and traffic condition from Waze. It is valuable to decouple the dynamic input from the static learning layer, then develop a flexible architecture that allows developers to save time in re-coding and wiring the solution.

## 6. Field Trip to Chariot: Going beyond the Science into the Real World

Like what David Silver, the lead designer of Google's Alpha Go, mentioned in his Ph.D. thesis, artificial intelligence (AI) solutions tend to work well in narrow domains, but fail in big world problems. The real world behaves in a form of Occam's razor, which means the simplest and clearest idea usually achieve good results. That said, being simple and intuitive is a key design principle when deploying AI solutions in the real world.

The design principle manifested in the choice of distributed RL. The distributed scheme breaks the problem to the most fundamental components; thus, it optimizes the problem at the least ambiguous and tangled condition. To achieve the maximum success, the theme needs to be carried out beyond the backend technical solution into the frontend experience that involves human operators. In the foreseeable future, most businesses will still require human staffs to execute computer-generated recommendations. For example, a trained dispatch staff needs to understand and validate the number of bikes to move recommended by the DiRL solution, then communicate the instructions to the logistic staff on the ground. Understanding how to design an intuitive solution that fits the business objectives and augment the performance of human staffs is a key to launching successful AI solutions in the real world.

The following section highlights key learnings from a field trip to Chariot's dispatch office in NYC. Chariot is a ride-sharing service that is backed by Ford Motor Company. Chariot provides cheap and flexible transportations to daily commuters using 14-seat vans. These vans operate on multiple fixed routes during rush hours. New routes can be proposed and designed or existing routes can be re-evaluated by commuters collaboratively; this is the on-demand aspect of the offering.

Although Chariot offers different services than CitiBike, the problems of manual asset allocation and heavy human interactions that the two companies face are the same. Hypothetically, the DiRL solution needs some refactoring in order to be deployed at Chariot, but the distributed RL architecture and TL scheme can be re-used without significant re-coding. The challenges are to find a viable value proposition of the RL solution and understand what frontend features may help the human staff the most, which many AI designers tend to neglect or de-prioritize.

The following photo shows the workstation of Peter Lau, who's an exceptional driver and dispatcher with over 10 years of experience, and the moment he was coordinating with van drivers to reroute due to traffic conditions. The field note highlights the environment, key tasks, and challenges of being a dispatcher.

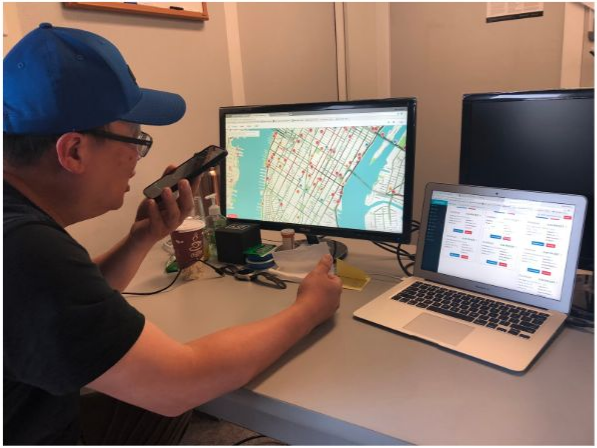

**Figure 9**: Field trip notes about the work of Chariot Dispatcher

According to the discussion with the Chariot team, which is summarized in Figure 9, the best value proposition of any AI solution is to help scale the fleet without adding dispatchers. An RL solution can be suitable given its capabilities to learn and improve in complex dynamics. However, having the right frontend features is imperative to the deployment success.

A useful "AI Assistant" to the dispatcher should have the following feature:
- Being able to digest all the real-time information and provide alerts or recommendations to help operators do their jobs better and faster
- Being able to automate predictable and quantifiable tasks, mainly re-routing and delay calculation in the case of Chariot; DiRL architecture can be modified to have van-agent pairs, instead of station-agent pairs
- Having an easy-to-use feedback interface to enhance AI performance and trust
- Having intuitive "nudges" as part of the workflow to encourage human interactions (e.g. Nudges based on Behavioural Science) (Abbeel, 2017)

## 7. Conclusion

In this project, I wanted to solve the bike rebalancing problem using an autonomous, scalable, and practical solution. With this in mind, I proposed and prototyped a distributed RL solution, DiRL, with Transfer Learning capabilities (same- and between-session). The DiRL allows multiple autonomous agents to move bikes within a network collectively, learn collaboratively, and improve without human intervention. In order to be practical, I took the opportunity to interview field operators at Chariot - a ride-sharing service that focuses on commuters - to understand how dispatchers work and how AI solutions should be deployed in the real world.

On the technical front, the DiRL shown promising performance in handling a network of bike stations. Although the solution still runs into scaling problems due to hardware constraints, it is exciting to see how the solution was able to improve the rebalancing performance by 350% autonomously and achieve over a 62.4% improvement in complete network success using Transfer Learning.

Conceptually, the field trip to Chariot provided ground truths to and directions on deploying AI systems. A successful AI solution must, first, align with business priorities, and second, provide concrete value by either allowing the current staff to do their jobs easier and/or helping business to scale without adding human overhead.

We are in a new wave of AI revolution with tremendous market traction and research investment. Commercializing AI research must balance innovation and practicality. I hope this project offer a playbook for students or industry professionals who are interested in bringing AI to the real world. We have an exciting journey ahead.

**Bibliography**

Abbeel, P. (2017, 08 08). Heroes of Deep Learning: Andrew Ng interviews Pieter Abbeel. (A. Ng, Interviewer)

Ali Yahya, A. L. (n.d.). *Collective Robot Reinforcement Learning with Distributed Asynchronous Guided Policy Search.* New York: Cornell University Library.

Barone, V. (2017, 06 11). *Citi Bike expansion making it harder to keep stations stocked*. From AM New York: https://www.amny.com/transit/citi-bike-expansion-making-it-harder-to-keep-stations-stocked-1.13730056

Etherington, D. (2017, 05 23). *Google's AlphaGo AI beats the world's best human Go player*. From Tech Crunch: https://techcrunch.com/2017/05/23/googles-alphago-ai-beats-the-worlds-best-human-go-player/

Hausman, K. (2018). *LEARNING AN EMBEDDING SPACE FOR TRANSFERABLE ROBOT SKILLS.* New York: ICLR 2018.

Jiang, Z. (2017). *A Deep Reinforcement Learning Framework for the Financial Portfolio Management Problem.* Suzhou: The Archive.

Kai Arulkumaran, M. P. (2017). *A Brief Survey of Deep Reinforcement Learning.* London: IEEE SIGNAL PROCESSING MAGAZINE.

Lazaric, A. (2013). *Transfer in Reinforcement Learning: a Framework and a Survey.* France: HAL.

Levine, S. (n.d.). *How Robots Can Acquire New Skills from Their Shared Experience.* San Fransisco: Google.

Lopes, D. M. (2015). *An Intelligent Bike-Sharing Rebalancing System.* Universidade de Coimbra. Coimbra: Universidade de Coimbra.

Mnih, V. (2015). Human-level control through deep reinforcement learning. *Nature*, 529-533.
Nourinejad, M. (2015). Vehicle relocation and staff rebalancing in one-way carsharing systems. *ScientDirect, 81*, 98-113.

Rahmatizadeh, R. (2018). *Vision-Based Multi-Task Manipulation for Inexpensive Robots Using End-To-End Learning from Demonstration.* New York: Cornell University Library.

Schuijbroek, J. (n.d.). *Inventory Rebalancing and Vehicle Routing in Bike Sharing Systems.* Carnegie Mellon University. Pittsburgh: Carnegie Mellon University.

Silva, F. L. (2016). Transfer Learning for Multiagent Reinforcement Learning Systems. *Twenty-Fifth International Joint Conference on Artificial Intelligence.* New York: Twenty-Fifth International Joint Conference on Artificial Intelligence.

Silver, D. (2017). Mastering the game of Go without human knowledge. *Nature*, 354-359.

Terekhova, M. (2017, 08 02). *JPMorgan takes AI use to the next level*. From Business Insider: https://www.businessinsider.com/jpmorgan-takes-ai-use-to-the-next-level-2017-8

Zhao, X. (n.d.). *Deep Reinforcement Learning for List-wise Recommendations.* Michigan : The Archive ARXIV.